\documentclass[11pt]{article} 
\usepackage{rldm,palatino}
\usepackage{graphicx}

\title{Option Discovery for Autonomous Generation \\ of Symbolic Knowledge}

\author{
Gabriele Sartor \\
University of Turin\\
Via Verdi, 8 \\
10124 Torino, Italy \\
\texttt{gabriele.sartor@unito.it} \\
\And
Davide Zollo \\
Roma Tre University \\
Via della Vasca Navale, 79\\
00146 Roma, Italy \\
\texttt{davidedezollo@gmail.com} \\
\And
Marta Cialdea Mayer \\
Roma Tre University \\
Via della Vasca Navale, 79 \\
00146 Roma, Italy \\
\texttt{cialdea@ing.uniroma3.it} \\
\And
Angelo Oddi \\
ISTC-CNR \\
Via San Martino della Battaglia, 44 \\
00185 Rome, Italy \\
\texttt{angelo.oddi@istc.cnr.it} \\
\And
Riccardo Rasconi \\
ISTC-CNR \\
Via San Martino della Battaglia, 44 \\
00185 Rome, Italy \\
\texttt{riccardo.rasconi@istc.cnr.it} \\
\And
Vieri Giuliano Santucci\\
ISTC-CNR \\
Via San Martino della Battaglia, 44 \\
00185 Rome, Italy \\
\texttt{vieri.santucci@istc.cnr.it} \\
}

\usepackage{algorithm}
\usepackage[noend]{algpseudocode}
\usepackage[autostyle]{csquotes}
\usepackage{caption}
\usepackage{subcaption}
\usepackage{amsmath}
\usepackage{xcolor}

\usepackage{wrapfig}

\begin{document}

\maketitle

\begin{abstract}
In this work we present an empirical study where we demonstrate the possibility of developing an artificial agent that is capable to autonomously explore an experimental scenario. During the exploration, the agent is able to discover and learn interesting options allowing to interact with the environment without any pre-assigned goal, then abstract and re-use the acquired knowledge to solve possible tasks assigned ex-post. We test the system in the so-called Treasure Game domain described in the recent literature and we empirically demonstrate that the discovered options can be abstracted in an probabilistic symbolic planning model (using the PPDDL language), which allowed the agent to generate symbolic plans to achieve extrinsic goals.
\end{abstract}


\acknowledgements{We are deeply indebted to George Konidaris and Steve James for making both the Skills to Symbols and the Treasure Game software available.}

\startmain

\section{Introduction}
In this work we want to show the possibility of developing an artificial system capable of "closing the loop" between low-level autonomous learning and high-level symbolic planning. In a previous work \cite{konidaris2018skills} the authors presented a promising approach, trying to combine learning in high-dimensional sensors and actuators spaces through the option framework with high-level, symbol-based, decision making through planning techniques. They proposed an algorithm able to abstract the knowledge of the system into symbols that can then be used to create sequences of operators to solve complex tasks in both simulated and real robotic scenarios. In particular, they assumed options were already given to the agent and then tested the capability of their algorithm to generate proper symbolic knowledge for a PDDL domain. 

The general problem addressed in this paper is therefore to develop a system that effectively starts from learning options that can then be abstracted through the algorithm in \cite{konidaris2018skills} and used to solve through planning a task in which it is necessary to perform a sequence of different actions. Option learning has been thoroughly studied in the literature: in particular, when assigned with a goal, a system can leverage on this assigned task to discover or generate sub-goals that can be learnt and encapsulated into options \cite{StollePrecup2002}. In the Treasure Game domain that we are considering here (see \cite{konidaris2018skills}), given the assigned task of reaching for the treasure, the different passages of climbing up and down a ladder, pulling levers, picking keys, etc., can be identified as components of the sequence of actions needed to achieve the final goal, and thus learnt through any learning algorithm, described as options and chunked together in the proper sequence. 
\begin{wrapfigure}{r}{0.47\textwidth}
    \centering
    \includegraphics[width=0.47\textwidth]{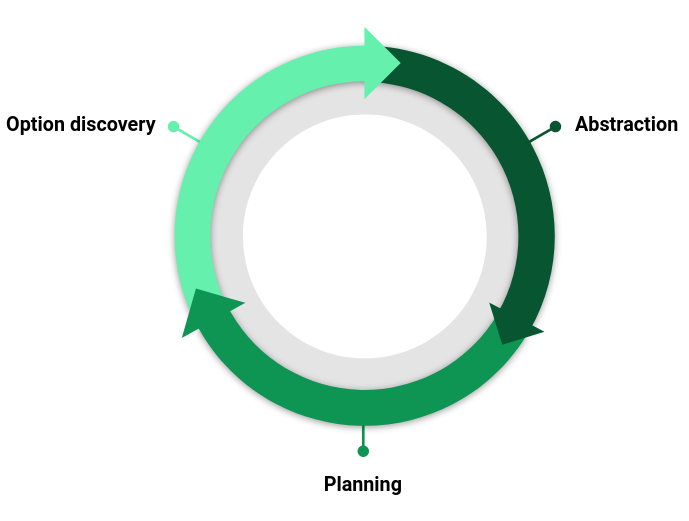}
    \caption{The proposed conceptual framework.}
    \label{fig:discovery_framework}
\end{wrapfigure}

However, the perspective we take in this work is that of an agent who has to learn autonomously to interact with the environment, without necessarily being aware of what tasks will be later assigned to it. We make this assumption because it forces us to develop versatile and adaptive agents that can be used in unknown and unstructured environments. 
In this perspective, the general problem described above becomes the more specific one of autonomously identifying which options to learn. 
Similarly to what done in \cite{SinghBartoChentanez2004}, we need to provide the agent with a general criterion by which it can identify states and/or events in the world that can be the target of specific options. 
Given that we do not know what task the agent will be asked to perform, we will leverage intrinsic motivations \cite{OudeyerKaplan2007} to identify potentially interesting states and use them to build options. 

The field of Intrinsically Motivated Open-ended Learning (IMOL, \cite{Santucci2020}) is showing promising results in the development of versatile and adaptive artificial agents. \textit{Intrinsic Motivations} (IMs, \cite{Oudeyer2007intrinsic}) are a class of self-generated signals that have been used to provide robots with an autonomous guidance for several different processes, from state-and-action space exploration \cite{Frank2014}, to the autonomous discovery, selection and learning of multiple goals \cite{Santucci2016,Colas2019}. In general, IMs guide the agent in the acquisition of new knowledge independently (or even in the absence) of any assigned task: this knowledge will then be available to the system to solve user-assigned tasks \cite{Seepanomwan2017} or as a scaffolding to acquire new knowledge in a cumulative fashion \cite{Santucci2010,Forestier2017} (similarly to what have been called curriculum learning \cite{Bengio2009}).

To connect this process of autonomous option discovery with the generation of high-level planning procedures, following our preliminary works \cite{OddiEtlAlECAI2020}, we adopt a hierarchical approach aimed at developing a robotic architecture capable of holding together the different mechanisms needed to close the loop between low and high level learning representations.    
In details, the idea depicted in Figure~ \ref{fig:discovery_framework} highlights the necessity of three different critical capabilities (or conceptual modules) that every robotic agent must have in order to operate autonomously.

The \textit{option discovery} module combines primitives to create options and learns precondition and effect models.
Then, these collected data are used by the \textit{abstraction} module to generate a PPDDL domain containing operators following the \textit{abstract sub-goal option} property.
Finally, the PPDDL description can be used by an off-the-shelf planner to reach any sub-goal which can be described with the available high-level symbols.
Potentially, the robotic agent can continue to execute these steps in a loop extending its knowledge and capabilities over time, exploitable by the human who can ask to reach a certain goal expressed in automatically generated symbols.
In the following sections we experimentally show that the discovered options can be abstracted in a probabilistic symbolic planning model (in PPDDL language), which allowed the agent to generate symbolic plans to achieve extrinsic goals.
\section{Option discovery}
      \label{sec:options}
\begin{wrapfigure}{r}{0.5\textwidth}
    \centering
    \includegraphics[width=8.5cm]{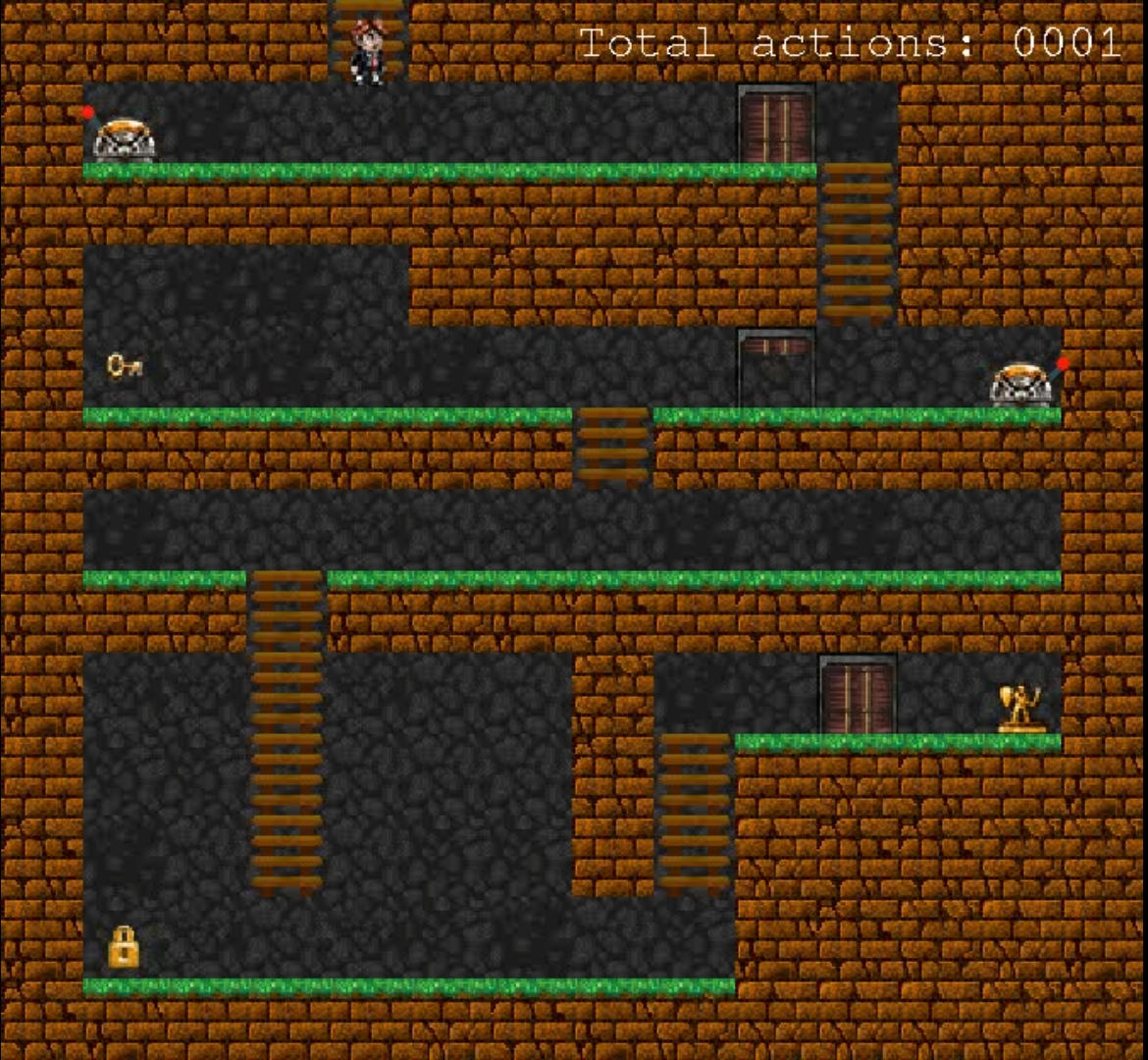}
    \caption{The Treasure Game configuration used for the experimental analysis.}
    \label{fig:treasure_Game}
\end{wrapfigure}
Options are temporally-extended actions defined as $o(I, \pi, \beta)$ \cite{Sutton1998}, in which $\pi$ is the policy executed, $I$ the set of states in which the policy can run and $\beta$ the termination condition of the option.
\begin{wrapfigure}{r}{0.45\textwidth}
    \centering
    \includegraphics[width=0.45\textwidth]{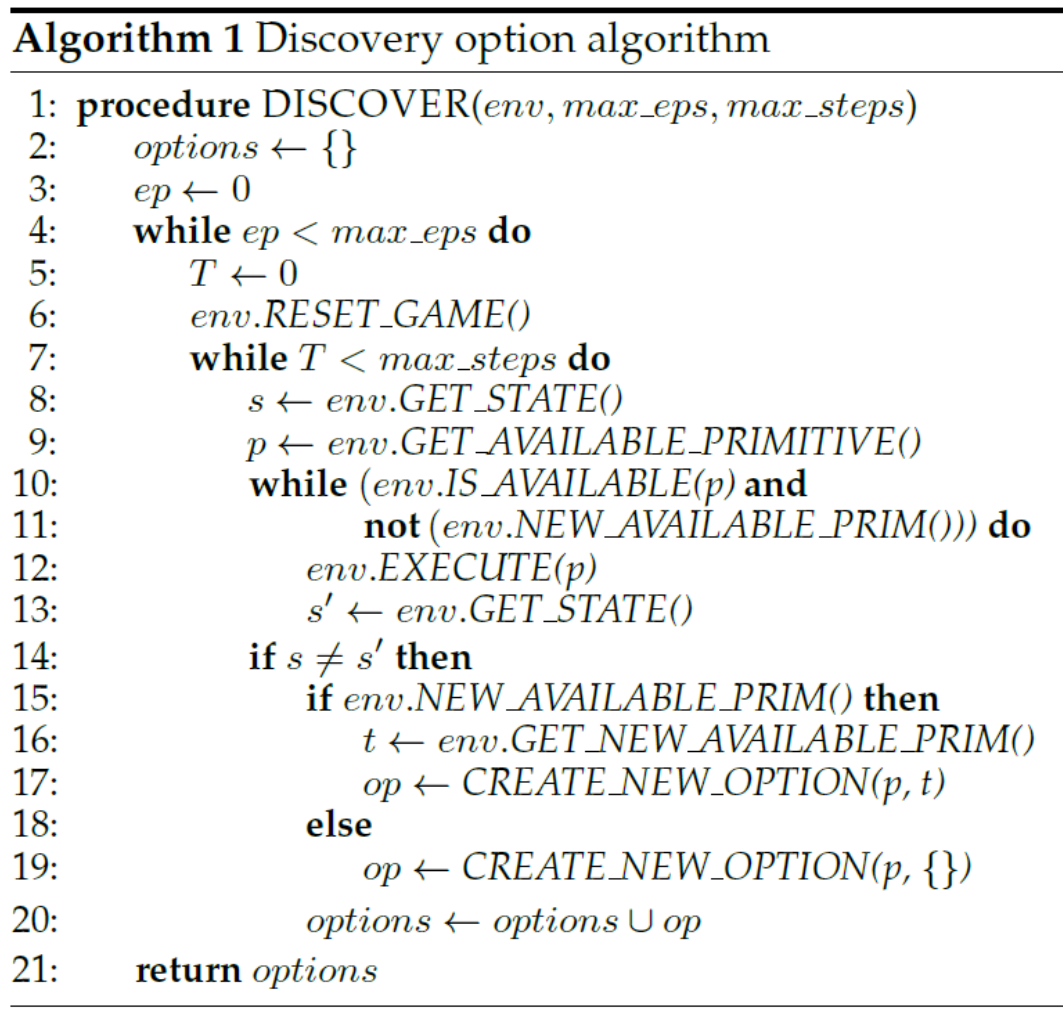}
    \caption{The algorithm discovering options in the environment.}
    \label{fig:rdml_algo}
\end{wrapfigure}
The option's framework revealed to be an effective tool to abstract actions and extend them with a temporal component.
The use of this kind of actions demonstrated to improve significantly the performances of model-based Reinforcement Learning compared to older models, such as one-step models in which the actions employed are the primitives of the agent \cite{Sutton:1999:MSF:319103.319108}.
Intuitively, these low-level single-step actions, or \textit{primitives}, can be  repeatedly exploited to create more complex behaviours.

In this section, we describe a possible way to discover and build a set of options from scratch. using the low-level actions available in the \textit{Treasure Game} environment (see Fig.~ \ref{fig:treasure_Game}). 
In such environment, an agent starts from its initial position (home), moves through corridors and climbs ladders over different floors, while interacting with a series of objects (e.g., keys, bolts, and levers) to the goal of reaching a treasure placed in the bottom-right corner and bringing it back home.

In order to build new behaviours, the agent can execute the following primitives: $1) go\_up$, $2) go\_down$, $3) go\_left$, $4) go\_right$, and $5)interact$,  respectively used to move the agent up, down, left or right by 2-4 pixels (the exact value is randomly selected with a uniform distribution) and to interact with the closest object.
In particular, the interaction with a lever changes the state (open/close) the doors associated to that lever (both on the same floor or on different floors) while the interaction with the key and/or the treasure simply collects the key and/or the treasure inside the agent's bag.
Once the key is collected, the interaction with the bolt unlocks the last door, thus granting the agent the access to the treasure.

In our experiment, primitives are used as building blocks in the construction of the option, participating to the definition of $\pi, I$ and $\beta$.
In more details, we create new options from scratch, considering a slightly different definition of option $o(p, t, I, \pi, \beta)$ made up of the following components: $p$, the primitive used by the execution of $\pi$; $t$, the primitive which, when available, stops the execution of $\pi$; $\pi$, the policy applied by the option, consisting in repeatedly executing $p$ until $t$ is available or $p$ can no longer be executed; $I$, the set of states from which $p$ can run; $\beta$, the termination condition of the action, corresponding to the availability of the primitive $t$ or to the impossibility of further executing $p$.
Consequently, this definition of option requires $p$, to describe the policy and where it can run, and $t$, to define the condition stopping its execution, maintaining its characteristic temporal abstraction.
For the sake of simplicity, the option's definition will follow the more compact syntax $o(p,t)$ in the remainder of the paper.
Algorithm~1 describes the process utilized to discover new options autonomously inside the simulated environment.
The procedure runs for a number of episodes $max\_eps$ and $max\_steps$ steps.
Until the maximum of steps of the current episode is not reached, the function keeps track of the starting state $s$ and randomly selects an available primitive $p$, such that $p$ can be executed in $s$ (lines $8$ and $9$).
Then, as long as $p$ is available and there is no new available primitives (lines $10$ and $11$), the option $p$ is executed, and the final state $s'$ of the current potential option is updated.
The function $\textit{NEW\_AVAILABLE\_PRIM}$ returns \textbf{True} when a primitive which was not previously executable becomes available while executing $p$; the function returns \textbf{False} in all the other cases.
For instance, if the agent finds out that there is a ladder over him while executing the $go\_right$ option, the primitive $go\_up$ gets available and the function return \textbf{True}.
In other words, $\textit{NEW\_AVAILABLE\_PRIM}$ detects the interesting event, thus implementing the surprise element that catches the agent's curiosity.
For this reason, the primitive representing the exact reverse with respect to the one currently being executed is not interesting for the agent, i.e., the agent will not get interested in the $go\_right$ primitive while executing $go\_left$.
The same treatment applied to the $(go\_left,go\_right)$ primitive pair is also used with the pair $(go\_up,go\_down)$.
When the stopping condition of the most inner while is verified and $s \neq s'$, a new option can be generated according to the following rationale.
In case the \textbf{while} exits because of the availability of a new primitive $t$ in the new state $s'$, a new option $o(p,t)$ is created (line $17$); otherwise, if the \textbf{while} exits because the primitive under execution is no longer available, a new option $o(p,\{\})$ is created, meaning \textit{"execute $p$ while it is possible"} (line $19$).
In either case, the created option $op$ is added to the list $options$ (line $20$), which is the output of the function.
In our test scenario, the algorithm generated 11 working options (see Section~\ref{sec:experiments}), suitable for solving the environment, and collected experience data to be abstracted in PPDDL \cite{YounesLittman2004}  format successively.
Consequently, as we introduced above, the agent performs two learning phases: the first, to generate options from scratch and creating a preliminary action abstraction, and the second, to produce a higher representation partitioning the options and highlighting their causal effects.
The latter phase, producing a symbolic representation suitable for planning, is analyzed in the next section.

\section{Empirical Analysis}
    \label{sec:experiments}

In this section we describe the results obtained from a preliminary empirical study, carried out by testing the Algorithm in Fig.~\ref{fig:rdml_algo} in the context of the Treasure Game domain \cite{konidaris2018skills}.
The algorithm was implemented in Python 3.7 under Linux Ubuntu 16.04 as an additional module of the \textit{Skill to Symbols} software, using the \textit{Treasure Game} Python package.
As previously stated, the Treasure Game domain defines an environment that can be explored by the agent by moving through corridors and doors, climbing stairs, interacting with handles (necessary to open/close the doors), bolts, keys (necessary to unlock the bolts) and a treasure.
In our experimentation, the agent starts endowed with no previous knowledge about the possible actions that can be executed in the environment; the agent is only aware of the basic motion primitives at his disposal, as described in Section~\ref{sec:options}.
The goal of the analysis is to assess the correctness, usability and quality of the abstract knowledge of the environment autonomously obtained by the agent.
The experiment starts by using the algorithm showed in Figure~\ref{fig:rdml_algo}, whose application endows the agent with the following set of learned options ($11$ in total):
\begin{equation}
    \begin{split}
    O = & \{(go\_up, \{\}), (go\_down, \{\}), (go\_left, \{\}), (go\_left, go\_up), (go\_left, go\_down), (go\_left, interact),\\ & (go\_right, \{\}), (go\_right, go\_up), (go\_right, go\_down), (go\_right, interact), (interact, \{\}) \}
    \end{split}
\end{equation}

The test has been run on an Intel I7, $3.4$~GHz machine, and the whole process took $30$ minutes.
All the options are expressed in the compact syntax $(p, t)$ described in Section~\ref{sec:options}, where \textit{p} represents the primitive action corresponding to the action's behavior, and \textit{t} represents the option's stop condition (i.e., the new primitive action discovered, or an empty set).
Once the set of learned options has been obtained, the test proceeds by applying the knowledge abstraction procedure described in \cite{konidaris2018skills}.
In our specific case, the procedure eventually generated a final PPDDL domain composed by a set of 1528 operators.
In order to empirically verify the correctness of the obtained PPDDL domain, we tested the domain with the off-the-shelf mGPT probabilistic planner \cite{BonetGeffner2005}.
The selected planning goal was to find the treasure, located in a hidden position of the environment (i.e., behind a locked door that could be opened only by operating on a bolt with a key) and bring it back to the agent's starting position, in the upper part of the Treasure Game environment.
The agent's initial position is by the small stairs located on the environment's $5^{th}$ floor (up left).
\begin{figure}
 \begin{center}
 \begin{verbatim}
1. go_down [to 5th floor]   2. go_left [to handle]      3. interact(handle) 
4. go_right [to wall]       5. go_down [to 4th floor]   6. go_right [to handle]
7. interact(handle)         8. go_left [to key]         9. interact(key)
10. go_right [to stairs]    11. go_down [to 3rd floor]  12. go_left [to stairs]
13. go_down [to 1st floor]  14. go_left [to bolt]       15. interact(bolt, key)
16. go_right [to wall]      17. go_up [to 2nd floor]    18. go_right [to treasure]
19. interact(treasure)      20. go_left [to stairs]     21. go_down [to 1st floor]
22. go_left [to bolt]       23. go_right [to stairs]    24. go_up [to 3rd floor]
25. go_right [to wall]      26. go_left [to stairs]     27. go_up [to 4th floor]
28. go_right [to handle]    29. interact(handle)        30. go_left [to stairs] 
31. go_up [to 5th floor]    32. go_left [to stairs]     33. go_up [home]
\end{verbatim}
\caption{Plan generated by the mGPT planner with our autonomously synthesized PPDDL domain.
\label{fig:generated_plan}}
\end{center}
\end{figure}
The symbolic plan depicted in Fig.~\ref{fig:generated_plan} was successfully generated and, as readily observable, reaches the goal that has been imposed (note that the PPDDL operators are named after their exact semantics manually, in order to facilitate their interpretation for the reader).
The previous analysis is still ongoing work, in this respect, there are at least three research lines to investigate.
The first line entails the study of different fine-tuning strategies of all the parameters utilized in the previously mentioned Machine Learning tools (such as DBSCAN, SVM, Kernel Density Estimator) involved in the knowledge-abstraction process. 
The second line is about analyzing the most efficient environment exploration strategy used to collect all the transition data that will be used for the classification tasks that are part of the abstraction procedure, as both the quantity and the quality of the collected data may be essential at this stage. 
The third one will be the exploration of innovative iterative procedures to incrementally refine \cite{Hayamizu_Amiri_Chandan_Takadama_Zhang_2021} the generated PPDDL model.

\bibliographystyle{splncs04}
\bibliography{main}

\begin{thebibliography}{10}
\providecommand{\url}[1]{\texttt{#1}}
\providecommand{\urlprefix}{URL }
\providecommand{\doi}[1]{https://doi.org/#1}

\bibitem{Bengio2009}
Bengio, Y., Louradour, J., Collobert, R., Weston, J.: Curriculum learning. In:
  Proceedings of the 26th annual international conference on machine learning.
  pp. 41--48 (2009)

\bibitem{BonetGeffner2005}
Bonet, B., Geffner, H.: Mgpt: A probabilistic planner based on heuristic
  search. J. Artif. Int. Res.  \textbf{24}(1),  933–944 (Dec 2005)

\bibitem{Colas2019}
Colas, C., Fournier, P., Chetouani, M., Sigaud, O., Oudeyer, P.Y.: Curious:
  intrinsically motivated modular multi-goal reinforcement learning. In:
  International conference on machine learning. pp. 1331--1340. PMLR (2019)

\bibitem{Forestier2017}
Forestier, S., Portelas, R., Mollard, Y., Oudeyer, P.Y.: Intrinsically
  motivated goal exploration processes with automatic curriculum learning.
  arXiv preprint arXiv:1708.02190  (2017)

\bibitem{Frank2014}
Frank, M., Leitner, J., Stollenga, M., F{\"o}rster, A., Schmidhuber, J.:
  Curiosity driven reinforcement learning for motion planning on humanoids.
  Frontiers in neurorobotics  \textbf{7}, ~25 (2014)

\bibitem{Hayamizu_Amiri_Chandan_Takadama_Zhang_2021}
Hayamizu, Y., Amiri, S., Chandan, K., Takadama, K., Zhang, S.: Guiding robot
  exploration in reinforcement learning via automated planning. Proceedings of
  the International Conference on Automated Planning and Scheduling
  \textbf{31}(1),  625--633 (May 2021),
  \url{https://ojs.aaai.org/index.php/ICAPS/article/view/16011}

\bibitem{konidaris2018skills}
Konidaris, G., Kaelbling, L.P., Lozano-Perez, T.: From skills to symbols:
  Learning symbolic representations for abstract high-level planning. Journal
  of Artificial Intelligence Research  \textbf{61},  215--289 (2018),
  \url{http://lis.csail.mit.edu/pubs/konidaris-jair18.pdf}

\bibitem{OddiEtlAlECAI2020}
Oddi, A., Rasconi, R., Santucci, V.G., Sartor, G., Cartoni, E., Mannella, F.,
  Baldassarre, G.: Integrating open-ended learning in the sense-plan-act robot
  control paradigm. In: ECAI 2020, the 24th European Conference on Artificial
  Intelligence (2020)

\bibitem{OudeyerKaplan2007}
Oudeyer, P.Y., Kaplan, F.: What is intrinsic motivation? a typology of
  computational approaches. Frontiers in Neurorobotics  \textbf{1}, ~6 (2009).
  \doi{10.3389/neuro.12.006.2007},
  \url{https://www.frontiersin.org/article/10.3389/neuro.12.006.2007}

\bibitem{Oudeyer2007intrinsic}
Oudeyer, P.Y., Kaplan, F., Hafner, V.: Intrinsic motivation systems for
  autonomous mental development. IEEE transactions on evolutionary computation
  \textbf{11}(2),  265--286 (2007)

\bibitem{Santucci2010}
Santucci, V.G., Baldassarre, G., Mirolli, M.: Biological cumulative learning
  through intrinsic motivations: a simulated robotic study on development of
  visually-guided reaching. In: Proceedings of the Tenth International
  Conference on Epigenetic Robotics (EpiRob2010. pp. 121--128 (2010)

\bibitem{Santucci2016}
Santucci, V.G., Baldassarre, G., Mirolli, M.: Grail: A goal-discovering robotic
  architecture for intrinsically-motivated learning. IEEE Transactions on
  Cognitive and Developmental Systems  \textbf{8}(3),  214--231 (2016)

\bibitem{Santucci2020}
Santucci, V.G., Oudeyer, P.Y., Barto, A., Baldassarre, G.: Intrinsically
  motivated open-ended learning in autonomous robots. Frontiers in
  neurorobotics  \textbf{13}, ~115 (2020)

\bibitem{Seepanomwan2017}
Seepanomwan, K., Santucci, V.G., Baldassarre, G.: Intrinsically motivated
  discovered outcomes boost user's goals achievement in a humanoid robot. In:
  2017 Joint IEEE International Conference on Development and Learning and
  Epigenetic Robotics (ICDL-EpiRob). pp. 178--183 (2017)

\bibitem{SinghBartoChentanez2004}
Singh, S., Barto, A.G., Chentanez, N.: Intrinsically motivated reinforcement
  learning. In: Proceedings of the 17th International Conference on Neural
  Information Processing Systems. p. 1281–1288. NIPS'04, MIT Press,
  Cambridge, MA, USA (2004)

\bibitem{StollePrecup2002}
Stolle, M., Precup, D.: Learning options in reinforcement learning. In: Koenig,
  S., Holte, R.C. (eds.) Abstraction, Reformulation, and Approximation. pp.
  212--223. Springer Berlin Heidelberg, Berlin, Heidelberg (2002)

\bibitem{Sutton1998}
Sutton, R.S., Barto, A.G.: Reinforcement learning: An introduction. MIT press
  (1998)

\bibitem{Sutton:1999:MSF:319103.319108}
Sutton, R.S., Precup, D., Singh, S.: Between mdps and semi-mdps: A framework
  for temporal abstraction in reinforcement learning. Artif. Intell.
  \textbf{112}(1-2),  181--211 (Aug 1999). \doi{10.1016/S0004-3702(99)00052-1},
  \url{http://dx.doi.org/10.1016/S0004-3702(99)00052-1}

\bibitem{YounesLittman2004}
Younes, H., Littman, M.: {PPDDL1.0: An Extension to PDDL for Expressiong
  Planning Domains with Probabilistic Effects}. Tech. rep., Carnegie Mellon
  University (2004), {CMU-CS-04-167}

\end{thebibliography}

\end{document}